\pdfoutput=1

\documentclass[11pt]{article}

\usepackage{acl}
\usepackage{times}
\usepackage{latexsym}

\usepackage[T1]{fontenc}

\usepackage[utf8]{inputenc}

\usepackage{hyperref}       
\usepackage{url}            
\usepackage{booktabs}       
\usepackage{amsfonts}       
\usepackage{nicefrac}       
\usepackage{microtype}      
\usepackage{xcolor}         
\usepackage{algorithm}      
\usepackage{blindtext}
\usepackage{graphicx}
\usepackage{multirow}
\usepackage{amsmath} 
\usepackage{enumitem}
\usepackage{colortbl}

\definecolor{lightblue}{rgb}{0.90, 0.90, 1.0}
\definecolor{lightgreen}{rgb}{0.6, 0.95, 0.6}

\definecolor{lightred}{rgb}{0.95, 0.6, 0.6}

\title{LLMs Meet Long Video: Advancing Long Video Question Answering \\with An Interactive Visual Adapter in LLMs}

\author{Yunxin Li$^{1}$, Xinyu Chen$^{1}$, Baotian Hu$^{1}$\thanks{~~~Corresponding author.}, Min Zhang$^{1}$\\
$^{1}$Harbin Institute of Technology, Shenzhen, China\\
\texttt{\{hubaotian, zhangmin2021\}}@hit.edu.cn
\\
\texttt{liyunxin987@163.com}
}

\begin{document}
\maketitle
\begin{abstract}

Long video understanding is a significant and ongoing challenge in the intersection of multimedia and artificial intelligence. Employing large language models (LLMs) for comprehending video becomes an emerging and promising method.
However, this approach incurs high computational costs due to the extensive array of video tokens, experiences reduced visual clarity as a consequence of token aggregation, and confronts challenges arising from irrelevant visual tokens while answering video-related questions.
To alleviate these issues, we present an \textit{Interactive Visual Adapter (IVA)} within LLMs, designed to enhance interaction with fine-grained visual elements.
Specifically, we first transform long videos into temporal video tokens via leveraging a visual encoder alongside a pretrained causal transformer, then feed them into LLMs with the video instructions. Subsequently, we integrated IVA, which contains a lightweight temporal frame selector and a spatial feature interactor, within the internal blocks of LLMs to capture instruction-aware and fine-grained visual signals. Consequently, the proposed video-LLM facilitates a comprehensive understanding of long video content through appropriate long video modelling and precise visual interactions. We conduct extensive experiments on nine video understanding benchmarks and experimental results show that our interactive visual adapter significantly improves the performance of video LLMs on long video QA tasks. Ablation studies further verify the effectiveness of IVA in long and short video understandings.

\end{abstract}


\section{Introduction}

The exponential advancement of the Internet and multimedia technologies has resulted in a significant surge in video content 
production by individuals and enterprises across various domains. The ability to interpret and extract meaningful content from videos is increasingly vital for meeting human demands and promoting the speed of information dissemination~\cite{tang2023video_survey}. Therefore, Video Question Answering~\cite{yu2019activityqa,seedBENCH,castro-etal-2022-in-the-wild} (Video QA), which allows users to ask about the content of videos through natural language and receive answers derived from their visual and auditory content, attracts tremendous research interest.
Recently, large language models (LLMs)~\cite{chatgpt,vicuna2023} have demonstrated exceptional efficacy in the domains of human-machine interaction and the handling of extensive contextual information. Capitalizing on these advancements, there is a burgeoning inclination towards integrating LLMs into the realm of video information processing. 
This approach primarily aims to enhance the interface between users and video content through intelligent question-and-answer (QA) sessions. 

The core of this innovation is a strategy that bridges the gap between the visual information in videos and the textual comprehension capabilities of LLMs. This is accomplished through a meticulously designed process that translates video data into a format comprehensible by LLMs, thereby facilitating an advanced question-answering system tailored for video content.
The process involves mapping video encoding into the language space of LLMs via a learnable visual mapping network~\cite{wu2023visual,li2023blip2,dai2023instructblip}. Essentially, the video is converted into ``video tokens'', which are then fed into the LLM along with textual tokens of natural language questions. Leveraging the vast knowledge storage and natural language processing prowess of LLMs, this approach effectively handles video QA tasks. For instance,
\citet{maaz2023video} performs spatial and temporal pooling for video tokens and feeds them into Vicuna~\cite{vicuna2023} to achieve the interaction between users and video content. \citet{zhang2023videollama} utilizes Q-former~\cite{li2023blip2} to extract question-relevant video tokens, which are then fed into LLaMA~\cite{gao2023llama} to generate the answer.

These LLMs-powered video understanding models~\cite{tang2023video_survey,song2023moviechat} mainly focus on short video modelling and have achieved a successful performance on short video captioning~\cite{damonlpsg2023videollama}, question-answering~\cite{jin2023video}, and summarization~\cite{tang2023video_survey}. However, the core challenges of video processing ~\cite{xu2023retrieval} stem from the need to efficiently model long video sequences and precisely respond to questions relevant to the video. Generally, using LLMs to handle long-form video often encounters the following hurdles: \textit{1) high computational costs from a multitude of video tokens; 2) reduced visual clarity as a consequence of token aggregation such as employing average or maximum representation pooling for visual frames; 3) irrelevant visual tokens leading to incorrect answers, notably when question-relevant information is embedded within long temporal cues.} Hence, previous models struggle to handle long-form videos owing to the constrained input capacity for video tokens and the challenge of distilling question-relevant, fine-grained visual features during generation.

To alleviate these issues, we present a long video comprehension method for LLMs, named \textit{Interactive Visual Adapter (IVA)} to achieve in-depth interactions between LLMs and video content. Specifically, we first use the pretrained visual encoder to obtain global and fine-grained frame representations. We construct the temporal video tokens by integrating the global features of frames with temporal video embeddings, which are obtained through a pretrained causal transformer. The whole set of temporal video tokens is fed into the LLM to attain a whole understanding of the video content. Additionally, we design a parameters-sharing Interactive Visual Adapter (IVA) that contains an instruction-aware temporal frames selector and a spatial feature interactor. The selector is used to obtain question-relevant frames based on contextual query embeddings and global encodings of videos. The selected frames are then fed into the spatial interactor to engage with the contextual query embeddings, in which fine-grained representations of frames are used. By doing so, LLMs could achieve in-depth interaction with video content by applying IVA between different layers.

To verify the effectiveness of our method, we conduct extensive experiments on four long video QA and five short video understanding benchmarks. Experimental results indicate that IVA is capable of achieving effective interactions between LLMs and long or short videos. 
Our contributions are summarized as follows:
\begin{itemize}
    \item We analyze the challenges of modelling long videos for LLMs and propose an interactive visual adapter for LLMs to handle long videos. It realizes the in-depth interaction between LLMs and long videos based on efficient video tokens and the IVA mechanism.
    \item The proposed IVA is capable of selecting relevant frames and interacting with their fine-grained spatial features through the internal selector and interactor, respectively. The IVA architecture is lightweight and designed to be shareable between layers of LLMs.
    \item Experimental results show that LLMs with IVA could achieve powerful performances in long video QA. Ablation studies underscore the critical role and effectiveness of IVA, confirming its significant contribution to enhanced performance.

\end{itemize}

\begin{figure*}[t]
    \centering
    \includegraphics[width=1.0\textwidth]{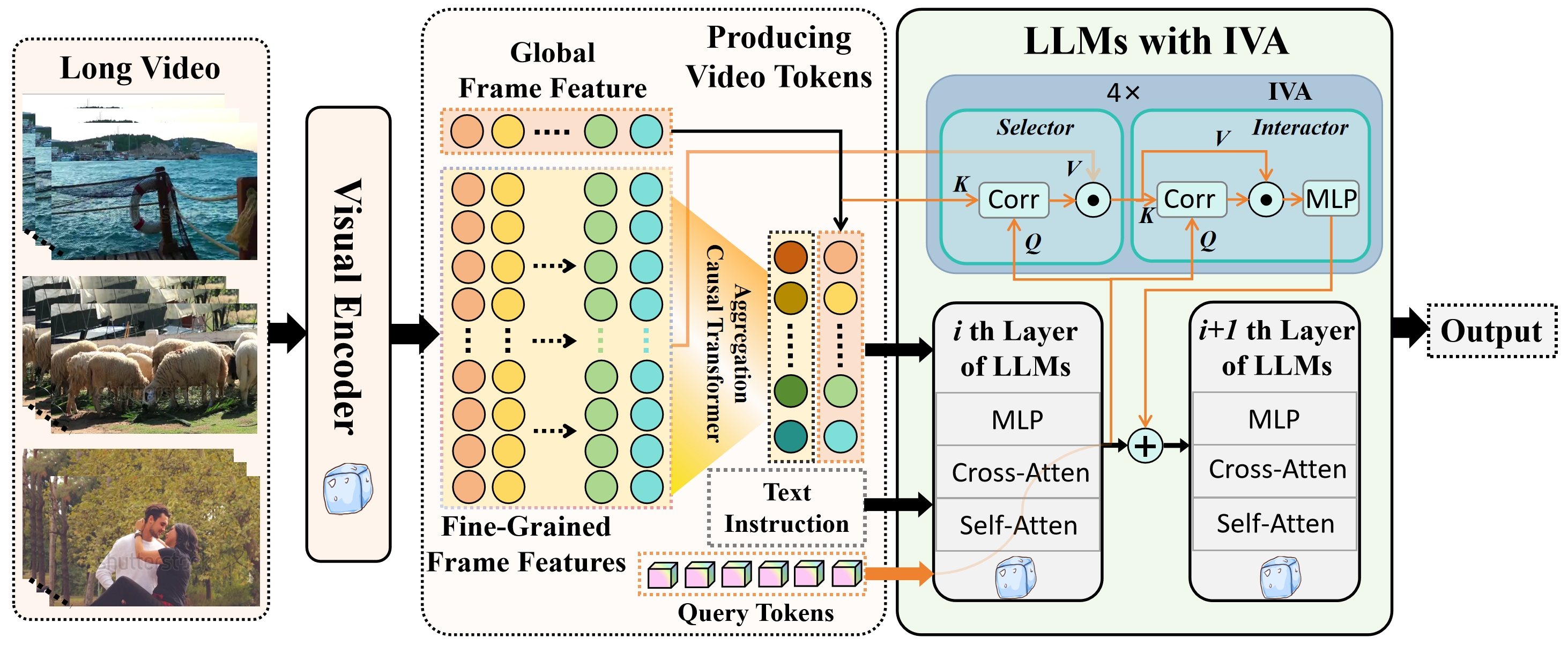}
    \caption{The overview of our framework employing LLMs to handle long video. While producing video tokens, we combine the global features and aggregated fine-grained features to represent a frame, allocating two tokens for each frame. The causal transformer is used to capture temporal relationships across frames and its output will be spliced with spatial feature sequence. The IVA will be inserted between blocks of LLMs to incorporate fine-grained visuals based on an understanding of the long video tokens, text instructions, and query tokens.}
    \label{fig:model}
\end{figure*}

\section{Related Work}

\textbf{Traditional Video Understanding Models}
The rapid development of deep learning methods possesses superior task-solving capabilities for video understanding. DeepVideo~\cite{KarpathyCVPR14} was the earliest method introducing a Convolutional Neural Network (CNN), for video understanding. Two-stream networks~\cite{feichtenhofer2016convolutional}, then integrating Convolutional Neural Networks (CNNs)~\cite{feichtenhofer2016convolutional} and Improved Dense Trajectories (IDT)~\cite{li2021vidtr}, enhanced motion analysis in video understanding. For long-form content, Long Short-Term Memory (LSTM)~\cite{yue2015beyond} networks were adopted, offering a robust solution for sequential data analysis over extended durations. Additionally, Temporal Segment Network (TSN)~\cite{wang2016temporal} advanced long-form video understanding by segmenting videos for individual analysis before aggregating insights, enabling more nuanced interpretation. Meanwhile, 3D networks started another branch by introducing 3D CNN to video understanding (C3D)~\cite{tran2015learning}. The introduction of Vision Transformers (ViT)~\cite{dosovitskiy2021an,arnab2021vivit,fan2021multiscale} promotes a series of prominent models 
Among the pioneering efforts in this self-supervised video training domain, VideoBERT~\cite{sun2019videobert} leverages the bidirectional language model BERT~\cite{kenton2019bert} for self-supervised learning from video-text data. This model, and others following the "pre-training and fine-tuning" paradigm, such as  ActBERT~\cite{zhu2020actbert}, SpatiotemporalMAE~\cite{feichtenhofer2022masked}, OmniMAE~\cite{girdhar2023omnimae}, showcase the diverse strategies developed to enhance video understanding. Notably, these models have set a foundation for advanced video-language models like  maskViT~\cite{gupta2022maskvit}, CLIP-ViP~\cite{xue2022clip}, LF-VILA~\cite{sun2022long}, further pushing the boundaries of what's achievable in action classification, video captioning, and beyond. The evolution from VideoBERT to more recent innovations like HiTeA~\cite{ye2023hitea}, and CHAMPAGNE~\cite{han2023champagne} underscores the rapid advancement in this field. 

\textbf{LLMs for Video Understanding}
The recent advancement in large language models (LLMs), pre-trained on expansive datasets, has ushered in groundbreaking capabilities in in-context learning \cite{zhang2023dnagpt} and long-form context modeling~\cite{lyu2023gpt}. This innovation has paved the way for integrating LLMs with computer vision technologies, exemplified by initiatives like Visual-ChatGPT~\cite{wu2023visual}. These models transcend traditional boundaries by calling vision model APIs~\cite{qin2023tool}, thereby addressing complex problems within the computer vision domain.
Integrating language models with video understanding technologies~\cite{maaz2023video,zhang2023llama,li2023videochat,xu2023retrieval,song2023moviechat} enhances multimodal understanding, facilitating sophisticated processing and interpretation of the intricate interplay between visual and textual data~\cite{ouyang2022training,liu2023visual,muennighoff2022crosslingual,li2023lmeye,zhu2023minigpt,zhang2023llama}. They leverage their extensive multimodal knowledge base and nuanced contextual understanding, mirroring a more human-like comprehension of video. Moreover, the exploration of LLMs in video understanding tasks~\cite{tang2023video_survey} represents a significant stride in analyzing and reasoning about visual data.

\section{Methodology}

\subsection{Overview}
Our work primarily introduces an interactive visual adapter for LLMs to handle long videos and answer relevant questions. The overview of workflow is shown in Figure~\ref{fig:model}. Specifically, given a video $V$, we first extract frames to obtain the whole sequence frame representations $\mathbf{h}_V = (\mathbf{h}_{I_1}, ..., \mathbf{h}_{I_k}, ..., \mathbf{h}_{I_N})$ via the pretrained image encoder, where $\mathbf{h}_{I_k} = (h_{g}^{I_k}, h_{1}^{I_k}..., h_{576}^{I_k})$ refers to the representations of $k$ th frame and $N$ is the total number of extracted frames. Then, we use a causal transformer to acquire temporal video embeddings from the aggregated spatial representation. The overall video tokens are formed by merging temporal video embeddings and global spatial features $[h_{g}^{I_1}, h_{g}^{I_k}, ..., h_{g}^{I_N}]$, where each frame is represented by two tokens. 
To enhance the capability of LLMs in leveraging fine-grained visual details from videos, we have developed an Interactive Visual Adapter (IVA) that is integrated into the blocks of LLM. This integration allows LLMs to comprehend the entirety of long videos through efficient video tokens while simultaneously capturing fine-grained visual information facilitated by the IVA.

\subsection{Producing Video Tokens}

We elaborate on the detailed process employed to produce efficient tokens for long videos, characterized by the extraction of one frame per second.
First, we use the self-weighted calculation on the fine-grained feature $\mathbf{h}_f^{k} = (h_1, ..., h_{576})$ of a frame ($k$ th) to obtain its overall representation, which will be fed into the following causal transformer.
This calculation process for the $k$ th frame is given in the following Eq.~\ref{eq1}:
\begin{equation}
\begin{array}{cc}
    s_f^{k} = Softmax( \mathbf{W}(\mathbf{h}_f^{k}) + \mathbf{b} ), \vspace{1.0ex} \\
    \mathbf{h}_t^{k} = s_f \mathbf{h}_f^{k},
    \end{array}
    \label{eq1}
\end{equation}
where $s_f \in \mathbf{R}^{1 \times 576}$ is the weight distribution and $\mathbf{W}$ and $\mathbf{b}$ are learnable parameters. Hence, we denote the obtained sequence-level frame representation as $\mathbf{h}_f = (\mathbf{h}_f^{1}, ..., \mathbf{h}_f^{N})$. 

\textbf{Causal Transformer} is employed to acquire the temporal video embeddings. Specifically, we use a four-layer transformer to facilitate interaction across frames, where a frame only attends to its previous ones. Take the first layer as an example, the specific operation of the causal transformer is presented in Eq.~\ref{eq2}:
\begin{equation}
    \begin{array}{cc}
        \mathbf{h}_s = SelfAtten (LayerN(\mathbf{h}_f), W^{Mask}) + \mathbf{h}_f \vspace{1.0ex}\\
        \mathbf{h}_s = LayerN(\mathbf{h}_s), \vspace{1.0ex} \\
        \mathbf{h}_o^{1} = MLP(\mathbf{h}_s)
    \end{array}
    \label{eq2}
\end{equation}
where $SelfAtten$ and $LayerN$ are the self-attention calculation and the feature normalization. The top output of the causal transformer will be projected into the language model by a linear layer, which is spliced with the global features $\mathbf{h}_{g}^{V} = (h_{g}^{I_1}, h_{g}^{I_k}, ..., h_{g}^{I_N})$. These global features of frames will be transferred into language models via a learnable MLP. We denote the final spliced feature to $\mathbf{h}_{V} = (h_{V}^{1}, h_{V}^{2}, ..., h_{V}^{2N})$.

\subsection{Interactive Visual Adapter}

After obtaining video tokens $\mathbf{h}_V$, and supposing that the textual embeddings of instruction are initiated to $\mathbf{h}_{T}$ via the frozen word embedding table of LLMs, we 
concatenate them into a single sequence and fed it into LLMs. Considering fine-grained visual details existing in long videos, we expect that LMMs are capable of capturing the specific fine-grained visual information based on the understanding of instructions and the whole video representations. Hence, we devise a lightweight interactive visual adapter (IVA) to enable LLMs to focus on instruction-relevant fine-grained visuals during content generation. 

Concretely, as the bottom part shown in Figure~\ref{fig:model}, we first introduce learnable dynamic tokens $\mathbf{h}_{D} = (h_1^{D}, ..., h_M^{D})$ as the query signals and integrate it at the end of the input token sequence. It aims to capture previous instruction and video information via the self-attention mechanism of LLMs, functioning as query tokens to engage with the fine-grained spatial features of videos. Suppose that the output of $i$ th layer of LLMs is $\mathbf{h}^{i}$. The specific calculation process of IVA between the $i$ and $i+1$ th layers of LLMs is shown in Eq.~\ref{eq3} and \ref{eq4} in order. Each layer of IVA consists of a selector and an interactor, which are capable of selecting relevant frames and capturing valuable fine-grained visual information. The operational process of the selector is described as follows:
\begin{equation}
    \begin{array}{cc}
         \mathbf{h}^{S}_{q} = W^{q} \mathbf{h}^{i}_{d} + b^{q}, \vspace{1.0ex} \\
          \mathbf{h}^{S}_{k} = W^{k} \mathbf{h}_{g}^{V}+ b^{k},\vspace{1.0ex}\\  
        \mathbf{M}^{S} = \mathbf{h}^{S}_{q} (\mathbf{h}^{S}_{k})^{T},\vspace{1.0ex}\\
        \mathbf{h}^{S} = Softmax(\mathbf{M}^{S} /\tau)Trans([\mathbf{h}_{I_1}^{f}, ..., \mathbf{h}_{I_N}^{f}]),    
    \end{array}
    \label{eq3}
\end{equation}
where $\mathbf{h}^{i}_{d}$ refers to the hidden states $\mathbf{h}^{i}$ associated with the indices of dynamic tokens. $ W^{q} $, $W^{k}$, $b^{q}$, and $b^{k}$ are learnable parameters. $\mathbf{M}^{S}$ signifies the distribution score on the frames, which represents the relevant attention distribution. $\tau$ is the hyperparameter, which is set to 0.5. ``Trans'' refers to the transportation of feature dimension. $[\mathbf{h}_{I_1}^{f}, ..., \mathbf{h}_{I_N}^{f}]$ represents the fine-grained features of the entire video. 
The output $\mathbf{h}^{S} \in \mathbf{R}^{b\times M \times 576 \times d_{S}}$ will be fed into the following interactor as the key value, where $d_{S}$ represents the dimension of the selector. 

For the interactor, the specific calculation progress could be given as Eq.~\ref{eq4}.
\begin{equation}
    \begin{array}{cc}
         \mathbf{h}^{I}_{q} = W^1 \mathbf{h}^{i}_{d} + b^1, \vspace{1.0ex} \\
          \mathbf{h}^{I}_{k} = W^2 \mathbf{h}^{S}+ b^2,\vspace{1.0ex}\\  
        \mathbf{M}^{I} = \mathbf{h}^{I}_{q} (\mathbf{h}^{I}_{k})^{T},\vspace{1.0ex}\\
        \mathbf{h}^{S}_c = Softmax(\mathbf{M}^{I})(W^3 \mathbf{h}^{S} + b^3), \vspace{1.0ex} \\
        \mathbf{h}^{S} = MLP (\mathbf{h}^{S}_c) + \mathbf{h}^{S}_c    
        \end{array}
    \label{eq4}
\end{equation}
where  $ W^{1} $, $W^{2}$,  $W^{3}$, $b^{1}$, $b^{2}$, and $b^{3}$ are learnable parameters. Overall, we use the same four-layer calculations of the above selector and interactor to facilitate that LLMs interact with fine-grained visual features. 

\subsection{Training}

\textbf{Stage 1: Pretraining}. To endow video tokens with meaningful representation, we first train the causal transformer, linear layers, and other learnable parameters during video tokens production, on massive video-caption pairs
from WebVid, a total of 703k video-caption pairs. We freeze the other parameters of the overall model during this process and do not introduce the IVA module.

\noindent \textbf{Stage 2: Video Instruction Tuning}. 
At this stage, the model is required to generate responses that align with various instructions. These instructions often involve complex visual comprehension and reasoning, rather than merely describing visual signals.
Note that the conversation data $[Q_1, A_1, ...,Q_r,A_r]$
consists of multiple rounds.
\begin{equation}
    X^r_T = 
\begin{cases} 
Q_{1}, & r = 1 \\
\text{Concat}(Q_1, A_1, ...,Q_r,A_r), & r > 1 
\end{cases}
\label{eq_f}
\end{equation}
where $r$ represents the round count. As shown in Eq. \ref{eq_f}, when $r>$ 1, we concatenate the conversations from all previous rounds with the current instruction as the input for this
round. The training objective remains the same as the previous stage. After this stage, the model can generate corresponding responses for various instructions.

\begin{table*}[t]
\renewcommand\arraystretch{1.05}
\centering
\scriptsize
    \scalebox{1.15}{
    \begin{tabular}{l| c c | c c | c c | c c}
        \toprule
        \multirow{2}{*}{\textbf{Method}} & \multicolumn{2}{|c|}{ActivityNet-QA} & \multicolumn{2}{|c|}{Social-IQ 2.0} & \multicolumn{2}{|c|}{LifeQA} & \multicolumn{2}{|c}{WildQA} \\
        \cline{2-9}
        & Accuracy & score & Accuracy & score & Accuracy & score & Accuracy & score \\
        \hline
        Video-LLaMA~\cite{zhang2023videollama} & 12.4 & 1.1 &  48.2 & 2.8 & 28.8 & 2.3 & \textbf{57.5} & 3.2 \\ 
        Video-Chat~\cite{2023videochat} & 26.5 & 2.2 & - & - & - & - & - & - \\
        LLaMA-Adapter~\cite{zhang2023llama} & 34.2 & 2.7 & - & - & - & - & - & - \\
        Video-ChatGPT~\cite{maaz2023video} & 35.2 & 2.7 & 51.6 & 3.2 & 31.2 & 2.5 & \underline{54.9} & 3.3\\
        \hline
        Baseline (w/o IVA) w/o Spatial Token & 38.4 & 3.0 & 49.8 & 3.1 & 26.3 & 2.2 & 51.4 & 3.0 \\
        Baseline (w/o IVA) w/o Causal Token & 38.2 & 2.9 & 49.9 & 3.1 & 30.6 & 2.4 & 51.7 & 3.1 \\
        Baseline (w/o IVA) & 40.8 & 3.0 & 53.0 & 3.3 & 30.9 & 2.5 & 54.4 & 3.2 \\
        \hline
        IVA (LQ=8, NI=8) & 41.6 & 3.0 & 54.0 & 3.6 & 46.5 & 2.8 & 51.2 & 3.1 \\
        \rowcolor{lightblue}
        IVA (LQ=16, NI=8) & 42.1 & 3.0 & \underline{64.9} & \underline{3.9} & 50.5 & 3.0 & 53.5 & \underline{3.2}\\
        IVA (LQ=32, NI=8) & 41.9 & 3.0 & 57.1 & 3.7 & \textbf{51.9} & \textbf{3.1} & 53.7 & 3.2\\
        IVA (LQ=16, NI=4) & 42.2 & 3.0 & 63.3  & 3.9 & 50.1 & 3.0 & 52.5 & 3.2\\
        IVA (LQ=16, NI=16) & \underline{42.3} & \underline{3.0} & 55.4 & 3.7 & 50.0 & 3.0 & 55.1 & \textbf{3.3}\\
        IVA (LQ=16, NI=8)-272K & \textbf{46.8} & \textbf{3.1} & \textbf{68.0} & \textbf{4.0} & \underline{48.1} & \underline{2.9} & 50.9 & 3.1\\
        \bottomrule
    \end{tabular}}
    \caption{\textbf{Comparison between different methods on 4 long video QA datasets.} LLM with IVA achieves the best performance on long videos compared to baselines and strong video LLMs. ``LQ'' refers to the length of query tokens and ``NI'' represents the number of interactions between LLMs and IVA. ``-272K'' indicates that we introduce additional training data of long video datasets like LifeQA and Social-IQ based on the original short video data.}
    \label{tab:long_video_result} 
\end{table*}

\section{Experiments}

\subsection{Data sets}

While training the causal transformer, we utilize 702 thousand video-text pairs derived from Valley 
\cite{luo2023valley}, sourced from WebVid \cite{Bain21}. During the instruction tuning stage, we collect instructional datasets from three sources: a 100K video-text instruction dataset from Video-ChatGPT \cite{Maaz2023VideoChatGPT}, a 36K short video-text instruction dataset from Valley-Instruct-73k \cite{luo2023valley}, and a 34K multiple-choice QA dataset from NExT-QA \cite{xiao2021next}. 
Additionally, we assess the generalization of IVA using long and short video benchmarks. Long video benchmarks typically are characterized by videos exceeding one minute in duration. We evaluated our model using four prominent long video evaluation benchmarks: ActivityNet-QA \cite{yu2019activityqa}, Social-IQ 2.0 \cite{siq2}, LifeQA \cite{castro-etal-2020-lifeqa}, WildQA \cite{castro-etal-2022-in-the-wild}.  For short video benchmarks, the duration of the videos is often measured in several seconds. We evaluate our model against three notable short video evaluation benchmarks: MSVD-QA \cite{xu2017video}, MSRVTT-QA \cite{xu2017video}, and SEED-Bench \cite{li2023seed}.

\subsection{Baselines}
We mainly compare our models with the following video LLMs that could be extended to handle long videos.
\textbf{Video-ChatGPT} \cite{Maaz2023VideoChatGPT} encodes frames independently and generates frame-level embeddings. Subsequently, it employs average pooling to transform these embeddings into both temporal and spatial features. These temporal and spatial features are then concatenated to derive video-level features and are fed into the LLM.
\textbf{Video-LLaMA} \cite{damonlpsg2023videollama} utilizes Vision-Language and Audio-Language to process video frames and audio signals separately. After fine-tuning on image instruction dataset and video instruction dataset, Video-LLaMA exhibited remarkable abilities in comprehending images and videos.
\textbf{Video-Chat} \cite{2023videochat} 
leverages perception tools to convert videos into textual descriptions in real-time, and employs a foundation model named InternVideo to encode videos into embeddings. These textual descriptions and video embeddings are then processed by an LLM for multimodal understanding.
\textbf{LLaMA-Adapter}~\cite{zhang2023llama} is a lightweight adapter injected into the attention calculation of LLM, which could be used to handle videos, text, and image tasks.


\begin{table*}[t]
\renewcommand\arraystretch{1.05}
\centering
\scriptsize
    \scalebox{1.20}{
    \begin{tabular}{l| c c | c c | c c c}
        \toprule
        \multirow{2}{*}{\textbf{Method}} & \multicolumn{2}{|c|}{MSVD-QA} & \multicolumn{2}{|c|}{MSRVTT-QA} & SEED\textsuperscript{AR} & SEED\textsuperscript{AP} & SEED\textsuperscript{PU} \\
        \cline{2-8}
        & Accuracy & score & Accuracy & score & Accuracy & Accuracy & Accuracy \\
        \hline
        Valley &- & - & - & - & 31.3 & 23.2 & 20.7 \\
        Video-LLaMA & 51.6 & 2.5 & 29.6 & 1.8 & - & - & -\\
        LLaMA-Adapter & 54.9 & 3.1 & 43.8 & 2.7 & - & - & -\\
        Video-Chat & 56.3 & 2.8 & 45.0 & 2.5 & 34.9 & \textbf{36.4} & 27.3\\
        Video-ChatGPT & \textbf{64.9} & \textbf{3.3} & 49.3 & 2.8 & 27.6 & 21.3 & 21.1 \\
        \hline
        Baseline (w/o IVA) & 54.5 & 3.2 & 49.6 & 2.9 & 22.5 & 23.5 & 24.8\\
        IVA (LQ=8, NI=8) & 53.2 & 3.2 & 47.6 & 2.9 & 32.0 & 31.8 & 27.5 \\
        \rowcolor{lightblue}
        IVA (LQ=16, NI=8) & 55.7 & 3.2 & 49.1 & 2.9 & \textbf{35.2} & 32.0 & \textbf{34.2} \\
        IVA (LQ=32, NI=8) & 53.0 & 3.2 & 47.2 & 2.9  & 32.2 & 32.1 & 28.8\\
        IVA (LQ=16, NI=4) & 55.0 & 3.2 & 47.8 & 2.9 & 32.5 & 31.7 & 26.0 \\
        IVA (LQ=16, NI=16) & 53.3 & 3.1 & 47.1 & 2.8 & 31.8 & 29.4 & 31.0 \\
        IVA (LQ=16, NI-8)-272K & 58.6 & 3.2 & \textbf{50.2} & \textbf{2.9} & 32.2 & 30.0 & 31.6 \\
        \bottomrule
    \end{tabular}}
    \caption{\textbf{Comparison between different methods on 5 zero-shot short video QA benchmarks.} Benchmark names are abbreviated due to space limits.
    MSVD-QA\cite{xu2017video};
    MSRVTT-QA\cite{xu2017video};
    SEED\textsuperscript{AR}: SEED-Bench Action Recognition\cite{li2023seed};
    SEED\textsuperscript{AP}: SEED-Bench Action Prediction\cite{li2023seed};
    SEED\textsuperscript{PU}: SEED-Bench  Procedure Understanding\cite{li2023seed}.
    }
    \label{tab:short_video_result} 
\end{table*}

\subsection{Evaluation Metrics}
For open-ended video QA tasks,  we employ ChatGPT-Assistant to evaluate the performance following Video-ChatGPT \cite{Maaz2023VideoChatGPT}. 
First, we input the question, the predicted answer, and the correct answer into ChatGPT. Second, we request ChatGPT to verify the accuracy of the predicted answer, expecting a binary response of 'yes' for correct predictions or 'no' for incorrect ones. Additionally, we require ChatGPT to rate the quality of the predicted answer on a scale from 0 to 5, where 5 indicates a perfect match. Finally, we determine the overall accuracy by counting the number of 'yes' responses and calculate the overall score by averaging all quality scores. This evaluation employs the "gpt-3.5-turbo" version of ChatGPT.

\subsection{Implementation Details}
We employ the AdamW optimizer~\cite{kingma2014adam}  in conjunction with a cosine learning rate scheduler to train our model. We first utilize 2 A100 GPUs to train visual-language MLP with 2 million image-text pairs with a global batch size of 256 and a base learning rate of 2e-4. Subsequently, we train the causal transformers using 703K video-text pairs data on the same two GPUs, employing a global batch size of 24 and a base learning rate of 3e-4. Transitioning to the video instruction tuning stage, we scale up to 8 A100 GPUs with a global batch size of 64. Here, we leverage LoRA to efficiently fine-tune the language model LLaMA. In our implementation, we set the rank to 128 and alpha to 256, maintaining a learning rate of 1e-4 for both LoRA and IVA parameters. Given the pretraining visual-language MLP and causal transformers, we adopt a smaller learning rate of 2e-5.

\subsection{Main Results}
We present the performance of the models on four long video QA benchmarks and five short video QA benchmarks. 
In long video QA benchmarks, our model achieved state-of-the-art (SOTA) results compared to the previous pure video LLMs, except WildQA. Especially on the LifeQA and Social-IQ 2.0 evaluation datasets, our model achieved significantly higher results, surpassing the previous SOTA accuracy by \textbf{18.0} and \textbf{7.4} percentage points, respectively. 
In short video QA benchmarks, our model also demonstrated strong capabilities across some evaluation datasets, especially in procedure understanding. We analyze the specific performance of WildQA in Appendix~\ref{wildqa_perform}.
Overall, IVA significantly enhances the capability of LLMs to analyze and interpret long videos, maintaining high-performance levels without compromising the understanding and reasoning abilities of short videos. 

\subsection{Ablation Study}


\noindent\textbf{Effect of IVA}.
From Tables \ref{tab:long_video_result} and \ref{tab:short_video_result}, introducing the IVA module significantly improves visual understanding in long video datasets (Social IQ2, LifeQA, ActivityNet-QA) and short video datasets. Notably, our model achieved over a 20\% improvement on LifeQA compared to the baseline, highlighting IVA's effectiveness. Comparing baselines without causal or spatial tokens confirms the efficiency of our video tokens production. The experimental results in Table \ref{tab:model_data} further suggest the beneficial impact of IVA when introducing more video instruction data. 

\begin{table}[t]
    \centering
    \renewcommand{\arraystretch}{1.10}
    \scriptsize
    \scalebox{0.8}{
    \begin{tabular}{lccccc}
        \toprule
        \textbf{Model} & \textbf{\#DataSize} & \textbf{SIQ2} & \textbf{LifeQA} & \textbf{WildQA} & \textbf{Avg.} \\
        \midrule
        Video-ChatGPT &  100k & 48.2/2.8 & 28.8/2.3 & 57.5/3.2 & 44.8 \\
        Video-LLaMA & 100k & 51.6/3.2	& 31.2/2.5	&54.9/3.3	& 45.9 \\
        Baseline(w/o IVA) & 100k & 53.0/3.3 &	30.9/2.5& 	54.4/3.2 & 46.1 \\
        IVA (LQ=16, NI=8) & 100k & 53.9/3.1 & 33.0/2.5	&57.4/3.3 & 48.1 $\uparrow$ \textcolor{blue}{2.0}\\
        \hline
        Baseline(w/o IVA)  & 170k & 56.7/3.2 & 31.7/2.3 & 52.2/3.1 & 46.9\\
        IVA (LQ=16, NI=8) & 170k & 64.9/3.9 & 50.5/3.0 & 53.5/3.2 & 56.3 $\uparrow$ \textcolor{blue}{6.4}\\
        \hline
        Baseline(w/o IVA)  & 272k & 59.3/3.3 & 34.5/2.4 & 50.5/3.1 & 48.1\\
        IVA (LQ=16, NI=8) & 272k & 68.0/4.0 & 48.1/2.9 & 50.9/3.1 & 55.7 $\uparrow$ \textcolor{blue}{7.6}\\
        \bottomrule
    \end{tabular}}
    \caption{Performance comparison when increasing the size of instruction data.}
    \label{tab:model_data}
\end{table}


\noindent\textbf{Length of Query Tokens}. Comparing the experimental results of IVA (LQ=8, NI=8) and IVA (LQ=16, NI=8) in Tables \ref{tab:long_video_result} and \ref{tab:short_video_result}, we observed a significant decrease in evaluation results across various benchmarks when reducing the length of query tokens (16 $\rightarrow$ 8). Regarding the comparison between IVA (LQ=16, NI=8) and IVA (LQ=32, NI=8) in long video benchmarks, we noted a slight decrease in performance on the first two benchmarks when increasing the length. However, while there was a slight improvement in LifeQA, it did not conclude an overall performance enhancement. In contrast, in the short video benchmarks, there was a downward trend in results across all benchmarks. Overall, increasing the length of query tokens may not lead to performance improvement. Moreover, reducing the length of query tokens may result in the loss of crucial visual information, consequently leading to performance degradation.

\begin{table}[t]
    \centering
    \scriptsize
    \scalebox{0.98}{
    \begin{tabular}{lcc}
        \toprule
        \textbf{Model} & \textbf{Total Second}$\downarrow$ & \textbf{Average Second}$\downarrow$ \\
        \midrule
        Video-ChatGPT & 96.83 & 0.9683  \\
        Video-LLaMA & 127.69 & 1.2769 \\
        Baseline(w/o IVA) & 89.18 & 0.8918 \\
        Baseline(IVA, LQ=16, NI=8) & 90.68 & 0.9068\\
        \bottomrule
    \end{tabular}}
    \caption{Comparison of model efficiency. All models are tested using the same 100 samples randomly selected from the evaluation set.}
    \label{tab:model_compute}
\end{table}

\begin{figure*}[h]
    \centering    \includegraphics[width=0.96\textwidth]{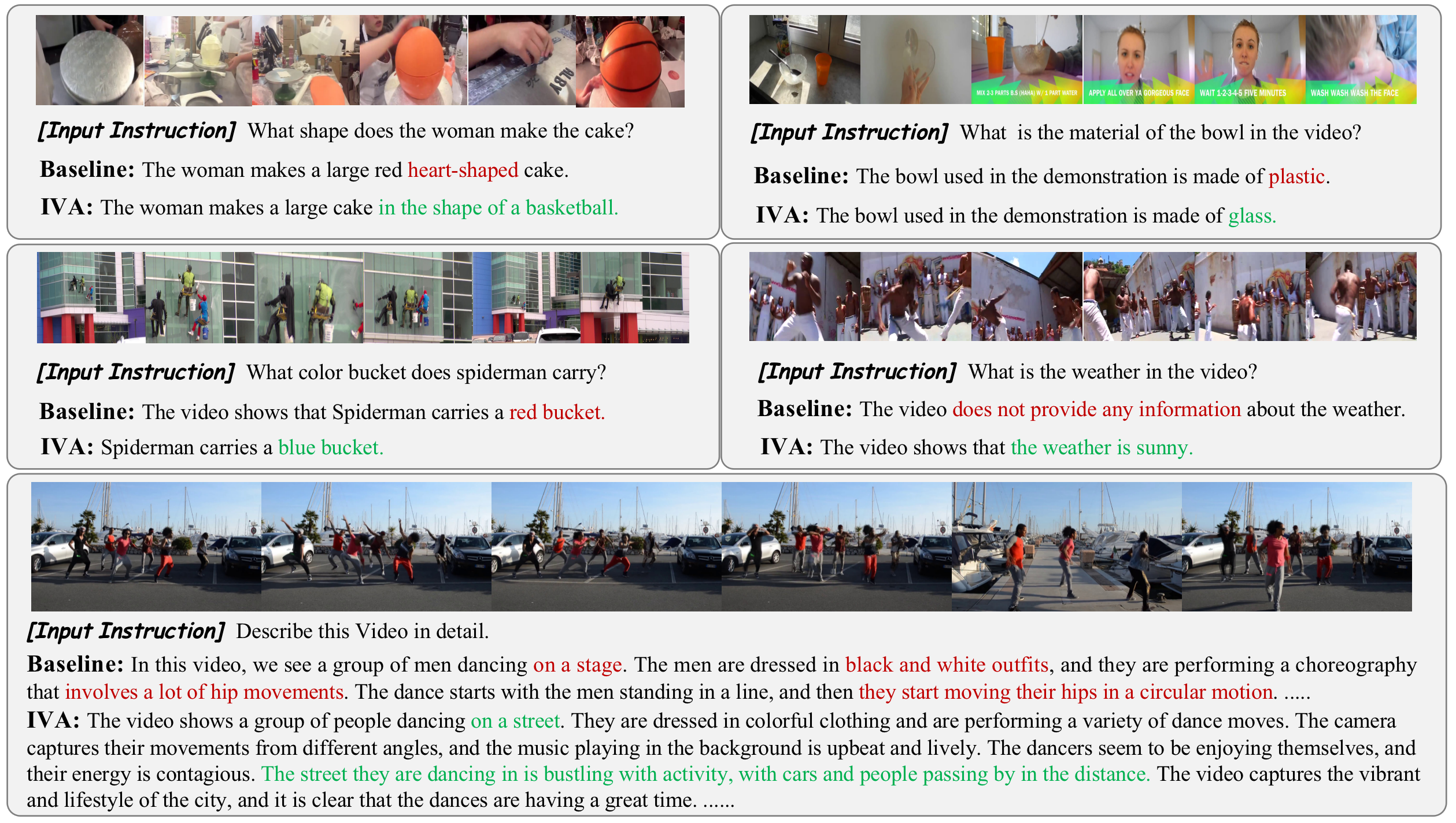}
    \caption{Five cases illustrate the comparative performances of our IVA Model and Baseline. \textcolor{red}{Red} and \textcolor{green!80}{green} words represent the inaccurate and accurate statements, respectively.}
    \label{fig:case_study_1}
\end{figure*}

\noindent\textbf{Impact of increasing long video instruction data}.
We explore enhancing the model's performance by introducing more long video data in Table~\ref{tab:model_data}. We use the training sets of Social IQ2 and LifeQA, the video instruction part of the MIMIC-IT dataset\cite{li2023mimicit}, along with the open-ended QA training data from NExT-QA, to the existing 170K training data, forming a new 272K training dataset. From the results of IVA(lQ=16, NI=8)-272K in Tables \ref{tab:long_video_result} and \ref{tab:short_video_result}, we observe a significant improvement in the Social IQ2 with the inclusion of more training data. However, there is little difference in the results on the remaining datasets, and in some cases, a certain degree of decline is observed. This may be attributed to the somewhat indiscriminate addition of new datasets, leading to a certain imbalance in the proportions of different data. The quality of the evaluation dataset will also affect the evaluation results, e.g., many questions in WildQA are typically wide-ranging and the criteria for answers can vary. We analyze this situation in Appendix~\ref{wildqa_perform}. Additionally, the training set of LifeQA only consists of 1,383 instances, which is relatively small in proportion to the total data, thus not providing sufficient improvement. While introducing more instruction data, IVA provides a bigger boost according to the results in Table~\ref{tab:model_data}.

\noindent\textbf{Number of interaction between IVA and  LLMs}.
We conduct experiments with both doubled and halved number of interaction layers. The detailed injection layers are shown in Appendix \ref{hyper-table}.
Upon analyzing the results of IVA(LQ=16, NI=8) and IVA(LQ=16, NI=4) in Tables~\ref{tab:long_video_result} and \ref{tab:short_video_result}, we observe that this reduction results in a significant decrease in its performance on most long video datasets, especially on the Action Prediction and Procedure Understanding of SEED-Bench. Moreover, the experimental results also indicate that increasing the number of layers (8 $\rightarrow$ 16) in the IVA interaction likewise causes a slight degradation in the model's performance. Given that there is no significant improvement observed when increasing the interaction times between IVA and LLMs, we set it to 8 as the standard for experimentation.

\noindent\textbf{Comparison of Computational Efficiency}.
We evaluate our baseline (w/o IVA), IVA (LQ=16, NI=8), and comparative models Video-ChatGPT and Video-LLaMA on 100 samples from the WildQA dataset in Table \ref{tab:model_compute}. All models used llama-7b or its variants with identical generation parameters. Results show:
1) Our method has the fastest inference speed, significantly surpassing Video-LLaMA.
2) With shared IVA and 8 interactions using query tokens of length 16, the increase in inference speed is marginal.
Our framework outperforms previous baselines in both inference efficiency and accuracy.

\subsection{Case Study}
We present four Video QA cases and one detailed description example in Figure \ref{fig:case_study_1}. 
Upon examining the initial two examples, we observe that the model augmented with IVA exhibits enhanced proficiency in recognizing particular actions associated with specific frames. In response to specific queries, it could discern objects such as the 'basketball-shaped cake', which solely appears towards the video's conclusion, and the 'glass bowl,' present solely in the video's opening segment. Furthermore, the fourth question-answering example illustrates that IVA augments the model's reasoning ability, enabling it to deduce the prevailing weather conditions based on the lighting conditions within the video. These indicate the effectiveness of IVA in incorporating fine-grained visuals of long videos.
Meanwhile, the bottom detailed description example reveals that when confronted with lengthy video descriptions, IVA could refine the perceptual acuity of LLMs, resulting in more precise recognition of elements such as the environment and colors.

\section{Conclusion}

In this study, we identify the principal obstacles in long video understanding and introduce an Interactive Visual Adapter (IVA) to facilitate dynamic interaction between LLMs and long videos. The IVA incorporates a selector module for identifying relevant temporal frames within long videos based on specific instructions and tokens, along with an interactor module that isolates detailed spatial visual features within long videos. The empirical results demonstrate that our IVA significantly improves LLMs' ability to comprehend and reason about long video content. 

\section*{Limitations}

Our work, while contributing insights into long video understanding and question-answering through employing LLMs, is subject to several limitations that warrant further investigation:
\begin{itemize}
\item \textbf{Optimization for Longer Videos}: Our current methodology demonstrates proficient performance in processing videos ranging from a few seconds to two minutes. However, the challenge of comprehensively understanding longer videos remains. Specifically, the optimization of video token length and the integration method of the Interactive Visual Adapter (IVA) within LLMs require further refinement to enhance their effectiveness and efficiency in handling extended content.

\item \textbf{Impact of Interaction Frequency and Query Token Length}: The stability of the IVA can be influenced by the frequency of interactions and the length of query tokens. These factors often occur in the development of multimodal large models, where a delicate balance must be struck between achieving high performance and maintaining operational efficiency, particularly in the context of long video interaction and encoding.

\item \textbf{Accuracy and Appropriateness of Generated Responses}: Another limitation is the potential for LLMs to generate responses that may be inaccurate, contain harmful content, or be factually incorrect. This issue stems from the inherent unpredictability in the response generation process of LLMs, underscoring the need for mechanisms that can ensure the reliability and appropriateness of the output.
\end{itemize}



\bibliography{anthology,custom}
\bibliographystyle{acl_natbib}

\appendix
\section{Inserting IVA in Different Layers}
\label{hyper-table}
\begin{table}[h]
\centering
\renewcommand\arraystretch{1.1}
\setlength{\tabcolsep}{8pt}
\small
\begin{tabular}{c | l c c}
\toprule
\textbf{NI} & \textbf{Corresponding Decoder Layers} \\
\midrule
4 & 0, 8, 16, 24 \\
8 & 0, 4, 8, 12, 16, 20, 24, 28 \\
16 & 0, 2, 4, 6, 8, 10, ..., 22, 24, 26, 28, 30 \\
\bottomrule
\end{tabular}
\caption{\textbf{Ablation Study on Injection Layers for IVA.} \textbf{NI}: Number of Inserting Layers. The incorporated inserting layers were positioned before the respective decoder layers. }
\label{tab:AblationStudyonNI}
\end{table}

In this section, we detail the methodology behind our ablation studies focusing on the variation in the Number of Injection Layers. Our experiments were structured around three different setups, where the injection layers were configured to be 4, 8, and 16 in number. To ensure a uniform distribution, these Injection Layers were interspersed throughout the decoder layers of the language model evenly. We utilized the Vicuna-7B model as our experimental framework, which is equipped with 32 decoder layers. The specific layers of the decoder that received the Injection Layers are outlined in Table~\ref{tab:AblationStudyonNI}, providing a clear reference to how the integration was achieved in each experimental setup

\section{Discussion of the lower performance of IVA on the WildQA dataset.}
\label{wildqa_perform}
We meticulously reviewed 652 question-and-answering pairs within the WildQA dataset, benchmarking various models against these inquiries. Of these, 319 questions were of the "what" variety, for instance, "What is the man doing?" or "What clothes is the woman wearing?" These questions are typically wide-ranging and the criteria for answers can vary, leading to inconsistencies in evaluation outcomes.

Additionally, a comparative analysis of "What"-specific accuracy versus overall sample accuracy revealed a direct correlation between IVA's performance on "what" questions and its overall accuracy. However, IVA prioritizes visual information comprehension and its responses depend on visual elements. After introducing more data for training, as the experimental results are shown in Tables \ref{tab:long_video_result} and \ref{tab:model_data}, improved performance across other datasets but not on WildQA further indicates the occasional disparity between its responses stemming mainly from video contents and the standard answers' format.

Example:
\textbf{Question}: What types of vehicles are being affected?
\textbf{Answer}: Cars, buses, motorcycles, mopeds.
\textbf{Prediction}: The video shows that cars and boats are being swept away by the flood water.

\begin{table}[h!]
\centering
\footnotesize
\scalebox{0.7}{
\begin{tabular}{lcc}
\hline
\textbf{Model} & \textbf{What-Acc (319 samples)} & \textbf{All-Acc (652 samples)} \\
\hline
Video-Chatgpt & 43.88 / 2.91 & 54.9 / 3.3 \\
Video-LLaMA & 43.26 / 2.79 & 57.5 / 3.3 \\
Baseline (w/o IVA) & 40.12 / 2.74 & 52.2 / 3.1 \\
IVA (LQ=16, NI=4) & 40.50 / 2.80 & 52.5 / 3.2 \\
IVA (LQ=16, NI=8) & 41.69 / 2.78 & 53.5 / 3.2 \\
IVA (LQ=16, NI=16) & 42.95 / 2.83 & 55.1 / 3.3 \\
IVA (LQ=16, NI=8)-272K & 39.23 / 2.73 & 50.9 / 3.1 \\
\hline
\end{tabular}}
\caption{Comparison of models on What-Acc and All-Acc metrics.}
\label{table:results}
\end{table}

\end{document}